\documentclass[conference]{IEEEtran}
\usepackage{ blindtext
           , graphicx
           , algorithm
           , algpseudocode
           }

\begin{document}

\title{Transfer learning approach for financial applications}

\author{
\IEEEauthorblockN{Cosmin Stamate}
\IEEEauthorblockA{Dept. of Computer science\\ and Information Systems\\
Birkbeck, University of London\\
London, United Kingdom\\
c.stamate@dcs.bbk.ac.uk}
\and
\IEEEauthorblockN{George D. Magoulas}
\IEEEauthorblockA{Dept. of Computer science\\ and Information Systems\\
Birkbeck, University of London\\
London, United Kingdom\\
gmagoulas@dcs.bbk.ac.uk}
\and
\IEEEauthorblockN{Michael S.C. Thomas}
\IEEEauthorblockA{Dept. of Psychological Sciences\\
Birkbeck, University of London\\
London, United Kingdom\\
m.thomas@psychology.bbk.ac.uk}
}

\maketitle

\begin{abstract}
Artificial neural networks learn how to solve new problems through a computationally intense and time consuming process. One way to reduce the amount of time required is to inject preexisting knowledge into the network. To make use of past knowledge, we can take advantage of techniques that \emph{transfer the knowledge} learned from one task, and reuse it on another (sometimes unrelated) task. In this paper we propose a novel selective breeding technique that extends the transfer learning with behavioural genetics approach proposed by Kohli, Magoulas and Thomas (2013), and evaluate its performance on financial data. Numerical evidence demonstrates the credibility of the new approach. We provide insights on the operation of transfer learning and highlight the benefits of using behavioural principles and selective breeding when tackling a set of diverse financial applications problems.
\end{abstract}


\section{Introduction}

It is fundamental for financial institutions to provide accurate risk assessments when evaluating new and/or existing customers (applicants). This risk assessment process should be continuous, evolve with the customer and/or institution requirements \cite{Khashman2010}. Most of the traditional assessment techniques that financial institutions use include but are not limited to: artificial neural networks (ANNs), support vector machines and complex tree structure predictors \cite{Crook2007}. It is wasteful to use these techniques in isolation as it does not exploit inter-departmental and/or inter-institutional sharing of \emph{valuable} information (knowledge). What we are proposing in this paper is a novel framework that has its roots in behavioural genetic studies, and provides a technique that facilitates the use of existing customer knowledge (from different departments or even between institutions) to improve the overall efficiency and speed of traditional customer evaluation techniques. This framework stems from previous work done by Kohli et al. \cite{kohlimt12, kohlimt13}, with the difference that it uses artificial breeding techniques between its populations of genetic algorithms to make this transfer of learning efficient on data with a financial aspect.

Genetic algorithms are artificial intelligence's search heuristic which can easily mirror the process of natural selection. Sub-branch of the vast evolutionary algorithms class that are inspired from the natural processes of evolution \cite{l.davis1991handbook-of-gen}. For genetic algorithms to behave in an evolutionary way we need to implement, all of the following evolutionary processes: selection, mutation, inheritance and crossover \cite{books/daglib/0092410}. In order to take full advantage of the power of evolutionary algorithms we need to use populations of abstractions (ANNs as individuals) which over iterations (time) will evolve towards better solutions; this will happen without the need of any external interference. Traditionally we start from random populations of individuals and iteratively improve on the quality of the population by evaluating each individuals fitness, selecting only the best performing and mating them. We stop the evolution when we reach a generation threshold or when we hit the fitness target for the population \cite{l.davis1991handbook-of-gen}.

One of the best performing classifiers of financial data used by financial institutions have been ANNs \cite{Khashman2010}. These have been tried and tested against statistically sound methods, and provide one of the most robust approaches to perform large-scale classification of financial data known to date \cite{Crook2007}. We have managed to use genetic algorithms (with novel selection and mating techniques) and traditional feed-forward ANNs and form a hybrid algorithm. This hybrid \emph{behavioural genetics} inspired algorithm is successful in performing transfer learning on heterogeneous tasks, and is also used to asses, to some degree, task similarity \cite{kohlimt13}.

In transfer learning the goal is to transfer useful knowledge from a source task (input) to a target task (output). We differentiate between two types of transfer paradigms: functional, where learning happens simultaneously in the source and in the target; and representational, where tasks are learned at different times \cite{pratt96, thrun:98a}. Most transfer learning research assumes a relationship between source and target tasks \cite{robins96, pratt96, thrun:98a}. Therefore the problem is that most of the transfer learning algorithms expect source and target tasks to be similar, otherwise they will not produce positive results \cite{kohlimt13, pratt96, thrun:98a}. Transfer learning between different tasks poses high risk as it might result in negative transfer, which damages learning capabilities. In traditional approaches, the risk of transfer having a negative impact on learning and performance is dictated by the degree of similarity between tasks \cite{kohlimt13, pratt96, thrun:98a}. We focus on the branch of transfer learning which deals with harnessing the ability to learn. We move the top performing lerners from generation to generation, until our learning models are very optimised for learning. By combining specific breeding techniques with the population based behavioural genetics framework we have achieved positive transfer learning on diverse financial tasks.

The paper is structured as follows. In the next section we give an overview of \emph{behavioural genetics} and explain the main ideas that underpin our approach. In Section~\ref{sec:methodology} the proposed methodology for transfer learning is derived and in Section~\ref{sec:datasets} the financial datasets are described. Section~\ref{sec:results} presents experimental results. The paper ends with concluding remarks in Section~\ref{sec:conclusion}.

\section{Overview of behavioural genetics}

Behavioural genetics studies the inheritance of traits between individuals of the same population, and then translates them into similarities and/or differences. Individual variation is separated in genetic and environmental components, which are examined for their influences on animal (including human) behaviour. In animal studies breeding and gene knockout techniques are commonly used to extract the required information. For humans the use of twin or adoption studies is more common. In this paper we have combined both human and animal studies by using both selective breeding and twin studies.

We have to stress the importance of the environment in individual development, this is simulated with the help of a filter applied to each training set. To get a better appreciation of the effect genetic and environmental influences have on network performance, we use the research on english past tense acquisition, with populations of identical and fraternal twins \cite{kohlimt12}. To provide better variation within populations, we use monozygotic (MZ, genetically identical twins, 100\% same genetic material) and dizygotic (DZ, genetically similar or fraternal twins, 50\% same genetic material), which in ANN terms translate to networks with identical hyper-parameters for MZ and 50\% identical hyper-parameters for DZ. Both type of networks (MZ or DZ) have random initial weights between their hidden nodes \cite{kohlimt13}.

Environmental influences are simulated by a unique filter which is applied to the training set for each individual of the population. The filtered training set resembles the notion of socio-economic-status (SES) \cite{kohlimt13}. Research on the effect of  SES on language acquisition shows that children with lower SES will perform significantly worse at the same task than children with a higher SES. This happens because of differences in the amount and quality of learning resources \cite{Thomas2013}. We ensured that the equal environment assumption was kept in all the simulated populations, which means that both MZ and DZ individual twins share the same environment \cite{kohlimt12, kohlimt13}. Splitting the populations between 50\% MZ and 50\% DZ twins, not only ensures good variability but proves helpful when attempting to learn multiple unrelated tasks \cite{kohlimt12}.

By incorporating and optimizing selective mating techniques into the behavioural genetics framework outlined above we have managed to focus and enhance the transfer of learning obtained on heterogeneous tasks \cite{kohlimt13}. We got the inspiration from the animal kingdom where certain animals have been mated (and still are now) for specific traits by humans, throughout history \cite{Jiang2013}.

\section{Proposed Transfer Learning Methodology}
\label{sec:methodology}

In this section we present the key components of our approach. This is based on a modification of the behavioural genetics population framework proposed by\cite{kohlimt13} that is combined with a new selective breading technique to enhance transfer of learning. Before we can use these algorithms we need to homogenise the feature dimensionality of our datasets, transforming them into environments with the same dimensionality for learning algorithms but keeping the internal distribution of the underlying problem the same. This is a requirement because we want our populations and individuals to be able to change environments without any external intervention on population and/or individual topology. Next we divide the data-sets into training, validation and testing, using a 60 - 20 - 20 distribution and we also split the data equally on the classes we have. This is useful not only from a machine learning context but it also mimics training with a curriculum \cite{journals/ftml/Bengio09}. It also ensures that training, validation and testing will have the same distribution of class examples and still have elements of randomness.

Algorithm~\ref{algo:hybrid} takes inspiration from the behavioural genetics framework developed by Kohli \cite{kohlimt13} but does not place any emphasis on heritability. A big part of Kohli's behavioural genetics framework is dedicated to heritability, which is abstracted, to some degree to task relatedness. We have used a more traditional approach to determine task relatedness, as outlined in the next section. SES has been randomly applied to each individual, from every population in the training phase, by removing vectors from the training set. SES is random and set between 0 and 40\%, the latter meaning that 40\% of the training set will be removed.

A novel selective breeding technique is presented in Algorithm~\ref{algo:breeding}. The selection part of this algorithm has no natural basis and has been developed in this way to fully exploit the variability and relationship between best and medium level performing twins. The best performing members of the population are selected for accuracy and medium level are selected to encourage population flexibility. Mixing mid level individuals with top performers is important for positive transfer of learning, otherwise, by choosing only top performers we have populations that are focused on optimal results and not on transfer. By mating top performers with medium level we obtain a good synergy between optimum results and transfer flexibility. The crossover and mutation part of the algorithm is the standard genetic algorithm random single point crossover and 0.1\% chance of random mutation.

\begin{algorithm}
\caption{Hybrid algorithm that exploits behavioural genetics intuitions in a transfer learning context}
\label{algo:hybrid}
\begin{algorithmic}[1]
\Require{$datasets \gets$ load all the preprocessed data-sets here }

\For{$dataset$ in $datasets$}
    \State{$populations \gets$ initialize 2 populations of the same size with random values (between specific intervals, predetermined by the distribution of the data-set) for the parameters of each individual}
    \While{$populations$ threshold is not reached} \Comment{we have used 20 generations as our threshold}
        \State{$brothers \gets emptyArray$}
        \For{$population$ in $populations$}
            \State{train($population$, $dataset$)} \Comment{only on training and validation sets, this also applies a random SES to the training set}
            \State{$fitness \gets$ assess($population$, $dataset$)} \Comment{on testing set only}
            \State{$leaders \gets$ selectForMating($population$, $fitness$)} \Comment{extract top and middle performers from the population}

            \State{$newPopulation \gets$ mate($leaders$)} \Comment{custom breeding method, outlined at algorithm~\ref{algo:breeding}}
            \State{$brothers \gets$ split($newPopulation$)} \Comment{split them into 2 equal populations (arrays of individuals) so that there are no identical twins in the same population}
        \EndFor
        \State{$populations \gets$ combine($brothers$)} \Comment{create 2 populations (arrays) from the 4 currently residing in $brothers$ by merging the populations that will not share any identical twins between them}
    \EndWhile
\EndFor
\end{algorithmic}
\end{algorithm}

\begin{algorithm}
\caption{Selective breeding algorithm for transfer of learning}
\label{algo:breeding}
\begin{algorithmic}[1]

\Require{$top \gets$ top performers from a population}
\Require{$mid \gets$ average performers from a population}

\State{$children \gets$} crossover($top$) \Comment{combine all the top performers in pairs and produce 4 offspring for each pair: 2 MZ twins, 2 DZ twins}
\State{$children \gets$} crossover($top, mid$) \Comment{combine middle and top performers in pairs and produce 4 offspring for each pair: 2 MZ twins, 2 DZ twins}
\State{$children \gets$} crossover($mid$) \Comment{combine all the middle performers in pairs and produce 4 offspring for each pair: 2 MZ twins, 2 DZ twins}

\end{algorithmic}
\end{algorithm}

\section{Datasets}
\label{sec:datasets}

All the sets of data used in this work have been selected from the UC Irvine Machine Learning Repository \cite{Lichman:2013} and are comparable with data published in \cite{TSAI2008, pany10, Khashman2010, Zhao2015}. Two of the three datasets are industry standards when it comes to assessing classifier accuracy on financial data \cite{TSAI2008, Khashman2010, Zhao2015}, and have been used to assess transfer of learning in \cite{pany10}. A short description of the datasets is available below and a comparison between characteristics in is shown in Table \ref{tab:datasets}.

\begin{itemize}
    \item \textbf{Statlog (Australian Credit Approval)}\footnote{\url{https://archive.ics.uci.edu/ml/datasets/Statlog+(Australian+Credit+Approval)}} (Australian) -- credit card applications with a good mix of attributes: continuous, small and large numbers of nominal values plus some missing values. This dataset was used in \cite{rossquinlan1993, quinlan1999simplifying, pany10, TSAI2008}.

    \item \textbf{Statlog (German Credit Data)}\footnote{\url{https://archive.ics.uci.edu/ml/datasets/Statlog+(German+Credit+Data)}} (German) -- credit data provided by Prof. Hofmann and the numerical dataset provided by Strathclyde University. Used in \cite{TSAI2008, pany10, Khashman2010, Zhao2015}

    \item \textbf{Banknote Authentication}\footnote{\url{https://archive.ics.uci.edu/ml/datasets/banknote+authentication}} (Banknote) -- data extracted from real images of forged banknotes, with the help of an industrial camera. Provided by Volker Lohweg (University of Applied Sciences, Ostwestfalen-Lippe). Used in \cite{Lohweg2013}.
\end{itemize}

As you can see two of the datasets are from the SatLog project. The SatLog project involves comparing the performances of machine learning, statistical, and ANNs algorithms on data sets from real-world industrial areas including medicine, finance, image analysis, and engineering design. The two datasets chosen by us from the SatLog project are popular in the machine learning and the financial world.

\begin{table}[h]
    \begin{center}
        \begin{tabular}{|l|cc|c|}
            \hline
            Datasets & Inputs & Instances & Attribute types \\
            \hline
            Australian & 14 & 690 & 6 numerical and 8 categorical \\
            German & 24 & 1000 & numerical \\
            Banknote & 5 & 1372 & numerical \\
            \hline
        \end{tabular}
    \end{center}
    \caption{Dimensionality of feature space with attribute types. All targets have 2 classes.}
    \label{tab:datasets}
\end{table}

The networks had the same number of hidden nodes for each task (the other intrinsic parameters were optimally selected), and the mean weight difference calculated between the various tasks, as can be seen in Table~\ref{tab: weightdiff}.

To assess task relatedness on the training, validation and testing sets of data for each task, we have used the mean difference between the weight space of best performing ANNs of the same size (same number of input, hidden and output nodes and layers) \cite{pratt96}. We have chosen a fixed 100 hidden nodes ANN, and with the rest of the hyper-parameters (learning rate, momentum and slope of logistic function) varying to perform optimally on each of the different tasks. After finding the best hyper-parameters for each of our 3 datasets, we get the mean of the norm of the difference of the weight vectors between our datasets. You can see the outcome in Table~\ref{tab: weightdiff}: the smaller the difference the more related the tasks are.

\begin{table}[h]
\begin{center}
\begin{tabular}{|l||ccc|}
\hline Datasets             & Australian    & German        & Banknote      \\
Australian                  & 0.0000        & 0.0426        & 0.0448        \\
German                      & 0.0426        & 0.0000        & 0.0401        \\
Banknote                    & 0.0448        & 0.0401        & 0.0000        \\
\hline
\end{tabular}
\end{center}
\caption{The mean of the norm of the difference of the weight vectors.}
\label{tab: weightdiff}
\end{table}

\section{Experiments and results}
\label{sec:results}

Before any of the experiments can commence we have to make sure that the data is in the format described in section~\ref{sec:methodology}. We need the data normalized and split into training, validation and testing sets.

We have two main sections for results, firstly we got the optimal neuro-computational calibrations for each dataset. This helped us encode each source task with its optimal ranges and it has enabled us to narrow down the initial search space. Although we can get the best set of hyper-parameters, we are not after best performance, what we are interested in is using intervals that center around the best performing hyper-parameters. This will ensure that transfer of learning can occur as the source networks will not be \emph{focused} only on the source task, and will allow a smooth transition of \emph{learning} from source to target(s). Sometimes the best hyper-parameter cannot be interpreted as the center of an interval, in that case we take the lowest most significant neuro-computational parameter as lower bound of the interval and the highest as the other. We have chosen only 4 hyper-parameters to optimise: number of hidden nodes, learning rate, momentum and slope of logistic function, but many more can be incorporated in this framework. You can see the calibration bounds of the datasets in Table~\ref{tab:calibration}.

\begin{table}[h]
\begin{center}
\begin{tabular}{|l||cccc|}
\hline Datasets             & Hidden nodes  & Learning rate & Momentum      & Logistic slope \\
Australian   & 15 to 50       & 0.01 to 0.2   & 0.01 to 5.1   & 1 to 4        \\
German           & 5 to 30       & 0.01 to 0.4   & 0.1 to 1.2    & 0.8 to 2.1     \\
Banknote     & 5 to 15       & 0.01 to 0.15  & 0.01 to 0.01  & 0.01 to 1.2    \\
\hline
\end{tabular}
\end{center}
\caption{Optimal source task calibrations.}
\label{tab:calibration}
\end{table}

After we have the source calibrations, we start using the two algorithms described in section~\ref{sec:methodology}. For the purposes of these experiments we have used populations of 1200 individuals (ANNs), each trained for 1000 epochs, without early stopping \cite{morgan:generalization}. The fitness criterion was overall miss-classification error on the test data set. We have started with two random (but within the bounds presented in table~\ref{tab:calibration}) populations. We select a source task, from the 3 available and train the 2 random populations on this chosen task. Furthermore we evolve these populations by following the steps outlined in Algorithm~\ref{algo:hybrid} and Algorithm~\ref{algo:breeding} for 20 generations. This produced highly optimised populations for the chosen source task, and populations that are flexible enough to transfer the acquired learning knowledge to aid learning of new tasks. The flexibility is possible because of the selective breeding pressures outlined in Algorithm~\ref{algo:breeding}. We take the 2 populations, breed them into one using the same technique as before (get top and middle performers and breed them) but without producing MZ and DZ twins. This produces only 2 non-identical offspring per pair and get a 1200 individual population, which is the population that best encodes the selected source task. We do the same for the remaining 2 source tasks, and have now one 1200 individuals population optimised for each of our tasks (datasets).

\begin{figure}[ht]
    \centering
    \includegraphics[scale=0.65]{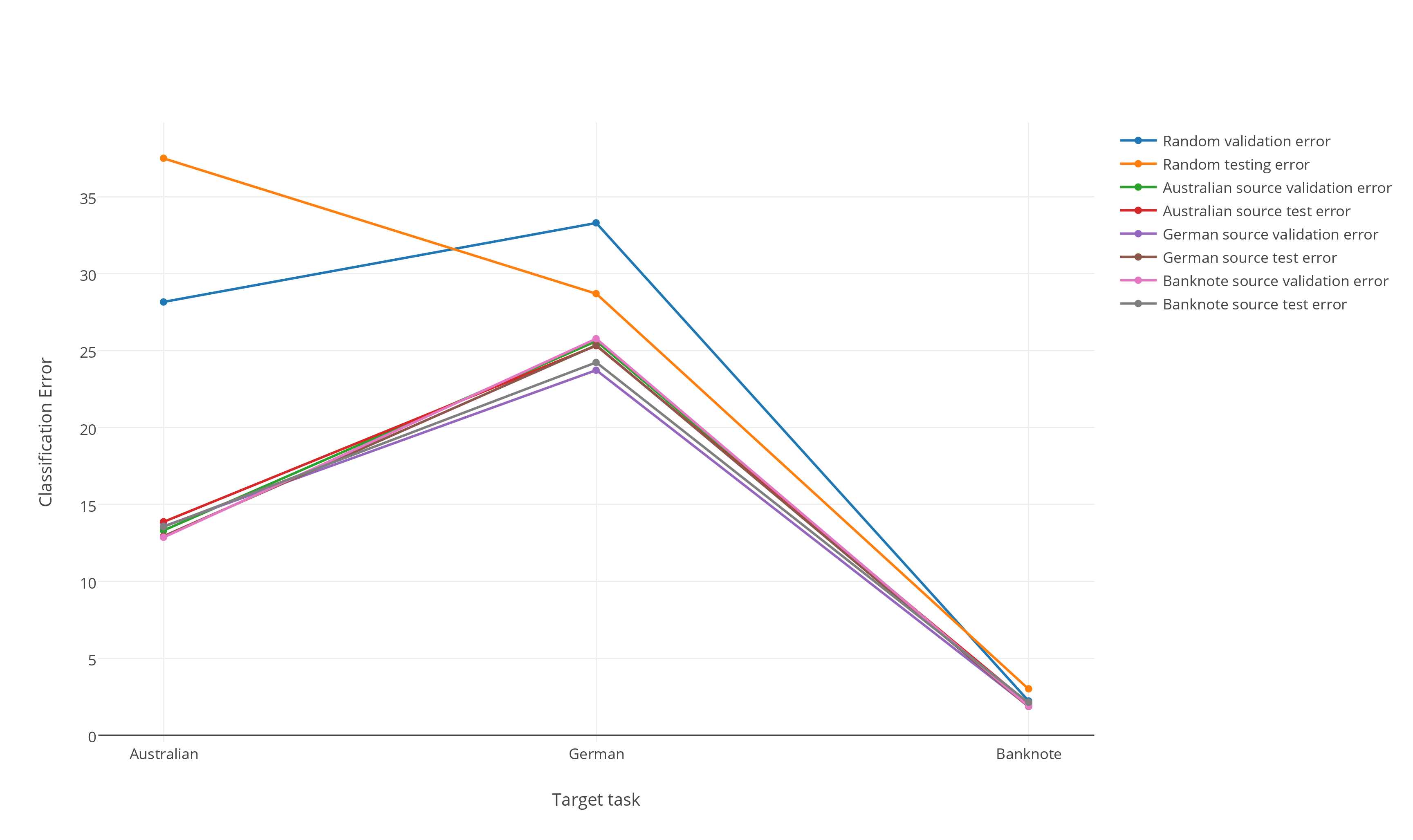}
    \includegraphics[scale=0.65]{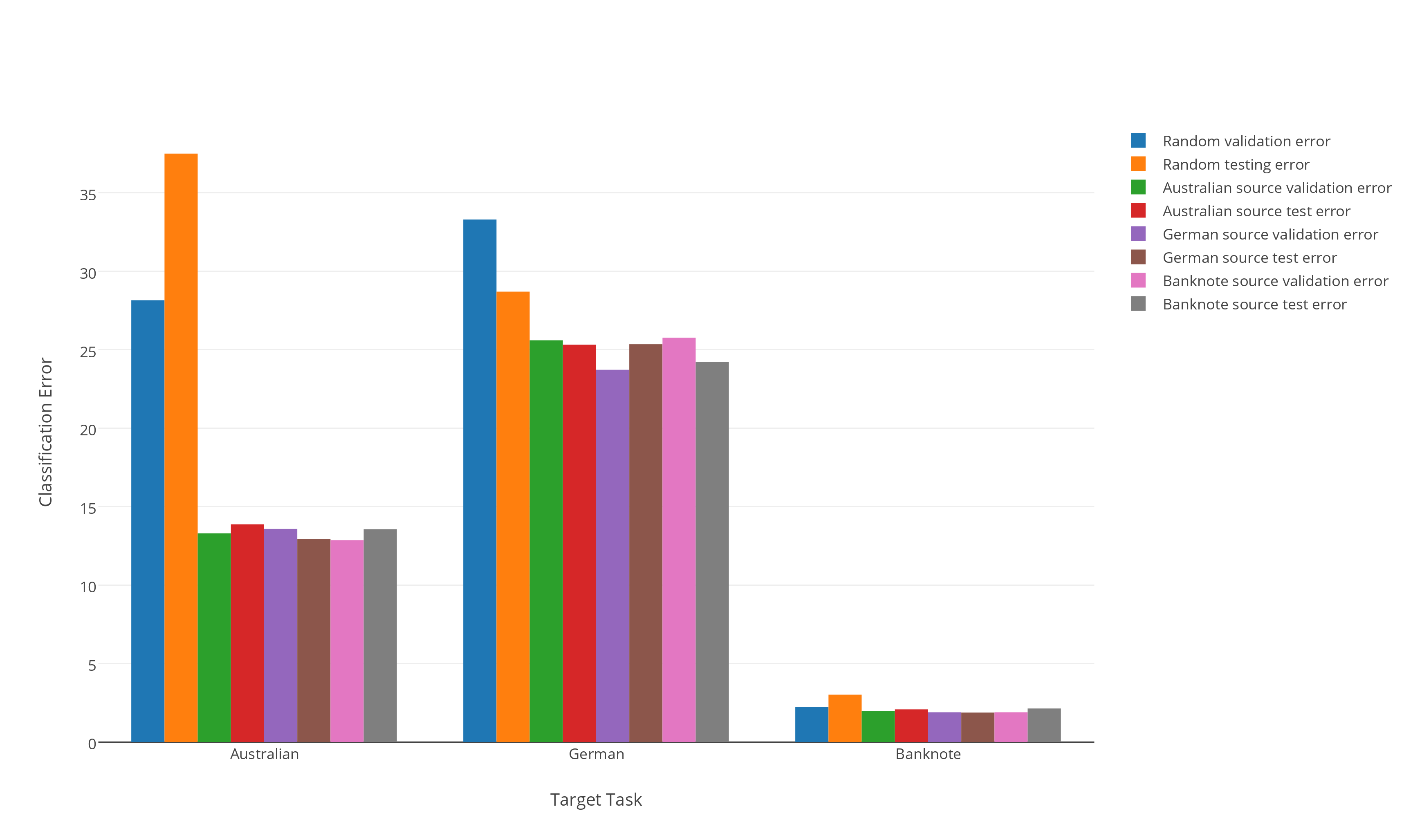}
    \caption{Benchmark between randomly initialised networks and our proposed transfer learning approach}
    \label{fig:benchmark}
\end{figure}

The optimised populations are then trained and tested on each of the tasks, and the results are presented in Tables~\ref{tab:australian}, \ref{tab:german} and \ref{tab:banknote}. The top part of each table represents the mean performance of the population on the validation data set of each target task, and the lower represents performance on the test set for each target task. When compared with results from literature \cite{TSAI2008, Khashman2010, Zhao2015}, this approach is not the best performing one. This is because we have focused our efforts in producing positive transfer of learning, without any negative impact on the overall process of learning. Whilst approaches in \cite{Khashman2010, TSAI2008, Zhao2015} were conceived to solve one of the problems outlined in the datasets in isolation. We have succeeded in solving all problems together and share knowledge between solutions, as one can see from the continuity of the results in Tables~\ref{tab:australian}, \ref{tab:german} and \ref{tab:banknote}. The continuity is not only in the overall classification error from source to target tasks but also in all the underlying parts that make that error. As one can see there is no spike or abrupt jump in values between all the true and false positives or negatives. When comparing our approach with a population (1200 individuals) of randomly initialised individuals (networks), we can see that our approach is considerably more accurate. This comparison becomes evident in Figure~\ref{fig:benchmark} with reinforcing testing and validation results presented in Table~\ref{tab:benchmark}.

\begin{table}[h]
\begin{center}
\begin{tabular}{|l||ccc|}
\hline Targets       & Australian    & German        & Banknote  \\
\hline
True positives                      & 330.75     & 137.5         & 737.25    \\
True negatives                      & 267.5      & 606.5         & 607.75    \\
False positives                      & 39.5       & 93.5          & 2.25      \\
False negatives                      & 52.25      & 162.5         & 24.75     \\
Precision               & 0.837      & 0.79          & 0.961     \\
Recall                  & 0.871      & 0.866         & 0.996     \\
Mean square error                   & 13.297     & 25.6          & 1.968     \\
\hline
True positives                      & 329.75     & 138.0         & 736.25    \\
True negatives                      & 264.5      & 608.75        & 607.25    \\
False positives                      & 42.5       & 91.25         & 2.75      \\
False negatives                      & 53.25      & 162.0         & 25.75     \\
Precision               & 0.833      & 0.790         & 0.959     \\
Recall                  & 0.862      & 0.870         & 0.995     \\
Mean square error                   & 13.877     & 25.325        & 2.077     \\
\hline
\end{tabular}
\end{center}
\caption{Evaluations on the validation (upper) and testing (lower) datasets for source task: \textbf{australian}.}
\label{tab:australian}
\end{table}

\begin{table}[h]
\begin{center}
\begin{tabular}{|l||ccc|}
\hline Targets       & Australian    & German        & Banknote  \\
\hline
True positives                     & 327.5      & 138.75        & 737.75    \\
True negatives                      & 268.75     & 624.0         & 608.25    \\
False positives                        & 38.25      & 76.0          & 1.75      \\
False negatives                      & 55.5       & 161.25        & 24.25     \\
Precision               & 0.829      & 0.795         & 0.962     \\
Recall                  & 0.875      & 0.891         & 0.997     \\
Mean square error                   & 13.587     & 23.725        & 1.895     \\
\hline
True positives                       & 330.75     & 127.0         & 737.5     \\
True negatives                         & 270.0      & 619.5         & 608.75    \\
False positives                       & 37.0       & 80.5          & 1.25      \\
False negatives                      & 52.25      & 173.0         & 24.5      \\
Precision               & 0.838      & 0.782         & 0.961     \\
Recall                  & 0.879      & 0.885         & 0.998     \\
Mean square error                   & 12.935     & 25.35         & 1.877     \\
\hline
\end{tabular}
\end{center}
\caption{Evaluations on the validation (upper) and testing (lower) datasets for source task: \textbf{german}.}
\label{tab:german}
\end{table}

\begin{table}[h]
\begin{center}
\begin{tabular}{|l||ccc|}
\hline Targets       & Australian    & German        & Banknote  \\
\hline
True positives                        & 337.5      & 111.75        & 738.75    \\
True negatives                      & 263.75     & 630.5         & 607.25    \\
False positives                        & 43.25      & 69.5          & 2.75      \\
False negatives                      & 45.5       & 188.25        & 23.25     \\
Precision               & 0.853      & 0.773         & 0.963     \\
Recall                  & 0.859      & 0.901         & 0.995     \\
Mean square error                   & 12.862     & 25.775        & 1.895     \\
\hline
True positives                        & 334.25     & 136.75        & 735.5     \\
True negatives                     & 262.25     & 621.0         & 607.25    \\
False positives                      & 44.75      & 79.0          & 2.75      \\
False negatives                      & 48.75      & 163.25        & 26.5      \\
Precision               & 0.843      & 0.793         & 0.958     \\
Recall                  & 0.854      & 0.887         & 0.995     \\
Mean square error                   & 13.550     & 24.225        & 2.132     \\
\hline
\end{tabular}
\end{center}
\caption{Evaluations on the validation (upper) and testing (lower) datasets for source task: \textbf{banknote}.}
\label{tab:banknote}
\end{table}

\begin{table}[h]
\begin{center}
\begin{tabular}{|l||ccc|}
\hline Targets       & Australian    & German        & Banknote \\
\hline
Validation error     & 28.155        & 33.3          & 2.22     \\
\hline
Testing error        & 37.5          & 28.7          & 3.01     \\
\hline
\end{tabular}
\end{center}
\caption{Classification error for randomly initialised networks}
\label{tab:benchmark}
\end{table}

\onecolumn
\twocolumn

\section{Conclusion}
\label{sec:conclusion}

Diverse financial data has been chosen as the topic of this work because financial institutions have and still are producing large amount of data on a daily basis. In contrast to approaches proposed in the literature where special methods are developed to solve a particular problem, we propose a transfer learning approach that exploits knowledge acquired in previous situations to learn a new problem. Approaches based on transfer learning are normally affected by negative transfer. The proposed framework based on behaviour genetics offers flexibility when learning different problems, alleviating the issue of negative transfer in the proposed datasets.  Furthermore, by using selective breeding, like humanity has done with animals for thousands of years, we have managed to improve overall accuracy and keep the learning flexibility. Novel selective breeding techniques injected in an already successful behavioural genetics computational framework, resulted in optimised positive transfer of learning. Although the results that we have reported are good from a classification and transfer point of view we could still improve this approach by utilizing more sophisticated selective breeding techniques and ensemble generation from populations based on individual class accuracy. In future work we intend to explore these methods in more depth.

\bibliographystyle{IEEEtran}
\bibliography{IEEEabrv,main}

\begin{thebibliography}{10}
\providecommand{\url}[1]{#1}
\csname url@samestyle\endcsname
\providecommand{\newblock}{\relax}
\providecommand{\bibinfo}[2]{#2}
\providecommand{\BIBentrySTDinterwordspacing}{\spaceskip=0pt\relax}
\providecommand{\BIBentryALTinterwordstretchfactor}{4}
\providecommand{\BIBentryALTinterwordspacing}{\spaceskip=\fontdimen2\font plus
\BIBentryALTinterwordstretchfactor\fontdimen3\font minus
  \fontdimen4\font\relax}
\providecommand{\BIBforeignlanguage}[2]{{%
\expandafter\ifx\csname l@#1\endcsname\relax
\typeout{** WARNING: IEEEtran.bst: No hyphenation pattern has been}%
\typeout{** loaded for the language `#1'. Using the pattern for}%
\typeout{** the default language instead.}%
\else
\language=\csname l@#1\endcsname
\fi
#2}}
\providecommand{\BIBdecl}{\relax}
\BIBdecl

\bibitem{Khashman2010}
\BIBentryALTinterwordspacing
A.~Khashman, ``Neural networks for credit risk evaluation: Investigation of
  different neural models and learning schemes,'' \emph{Expert Systems with
  Applications}, vol.~37, no.~9, pp. 6233--6239, sep 2010. [Online]. Available:
  \url{http://dx.doi.org/10.1016/j.eswa.2010.02.101}
\BIBentrySTDinterwordspacing

\bibitem{Crook2007}
\BIBentryALTinterwordspacing
J.~N. Crook, D.~B. Edelman, and L.~C. Thomas, ``Recent developments in consumer
  credit risk assessment,'' \emph{European Journal of Operational Research},
  vol. 183, no.~3, pp. 1447--1465, dec 2007. [Online]. Available:
  \url{http://dx.doi.org/10.1016/j.ejor.2006.09.100}
\BIBentrySTDinterwordspacing

\bibitem{kohlimt12}
M.~Kohli, G.~D. Magoulas, and M.~S.~C. Thomas, ``Hybrid computational model for
  producing english past tense verbs,'' in \emph{EANN}, ser. Communications in
  Computer and Information Science, C.~Jayne, S.~Yue, and L.~S. Iliadis, Eds.,
  vol. 311.\hskip 1em plus 0.5em minus 0.4em\relax Springer, 2012, pp.
  315--324.

\bibitem{kohlimt13}
------, ``Transfer learning across heterogeneous tasks using behavioural
  genetic principles,'' in \emph{UKCI}.\hskip 1em plus 0.5em minus 0.4em\relax
  IEEE, 2013, pp. 151--158.

\bibitem{l.davis1991handbook-of-gen}
L.~Davis, Ed., \emph{Handbook of Genetic Algorithms}.\hskip 1em plus 0.5em
  minus 0.4em\relax Van Nostrand Reinhold, 1991.

\bibitem{books/daglib/0092410}
T.~Back, \emph{Evolutionary algorithms in theory and practice - evolution
  strategies, evolutionary programming, genetic algorithms.}\hskip 1em plus
  0.5em minus 0.4em\relax Oxford University Press, 1996.

\bibitem{pratt96}
L.~Y. Pratt, ``A survey of transfer between connectionist networks,''
  \emph{Connect. Sci.}, vol.~8, no.~2, pp. 163--184, 1996.

\bibitem{thrun:98a}
S.~Thrun and L.~Pratt, \emph{Learning to Learn}.\hskip 1em plus 0.5em minus
  0.4em\relax Kluwer Academic Publishers, 1998.

\bibitem{robins96}
A.~Robins, ``Transfer in cognition,'' \emph{Connect. Sci.}, vol.~8, no.~2, pp.
  185--204, 1996.

\bibitem{Thomas2013}
\BIBentryALTinterwordspacing
M.~S.~C. Thomas, N.~A. Forrester, and A.~Ronald, ``Modeling socioeconomic
  status effects on language development.'' \emph{Developmental Psychology},
  vol.~49, no.~12, pp. 2325--2343, 2013. [Online]. Available:
  \url{http://dx.doi.org/10.1037/a0032301}
\BIBentrySTDinterwordspacing

\bibitem{Jiang2013}
\BIBentryALTinterwordspacing
Y.~Jiang, D.~I. Bolnick, and M.~Kirkpatrick, ``Assortative mating in animals,''
  \emph{The American Naturalist}, vol. 181, no.~6, pp. E125--E138, jun 2013.
  [Online]. Available: \url{http://dx.doi.org/10.1086/670160}
\BIBentrySTDinterwordspacing

\bibitem{journals/ftml/Bengio09}
\BIBentryALTinterwordspacing
Y.~Bengio, ``Learning deep architectures for ai.'' \emph{Foundations and Trends
  in Machine Learning}, vol.~2, no.~1, pp. 1--127, 2009. [Online]. Available:
  \url{http://dblp.uni-trier.de/db/journals/ftml/ftml2.html#Bengio09}
\BIBentrySTDinterwordspacing

\bibitem{Lichman:2013}
\BIBentryALTinterwordspacing
M.~Lichman, ``{UCI} machine learning repository,'' 2013. [Online]. Available:
  \url{http://archive.ics.uci.edu/ml}
\BIBentrySTDinterwordspacing

\bibitem{TSAI2008}
\BIBentryALTinterwordspacing
C.~TSAI and J.~WU, ``Using neural network ensembles for bankruptcy prediction
  and credit scoring,'' \emph{Expert Systems with Applications}, vol.~34,
  no.~4, pp. 2639--2649, may 2008. [Online]. Available:
  \url{http://dx.doi.org/10.1016/j.eswa.2007.05.019}
\BIBentrySTDinterwordspacing

\bibitem{pany10}
S.~J. Pan and Q.~Yang, ``A survey on transfer learning,'' \emph{IEEE Trans.
  Knowl. Data Eng.}, vol.~22, no.~10, pp. 1345--1359, 2010.

\bibitem{Zhao2015}
\BIBentryALTinterwordspacing
Z.~Zhao, S.~Xu, B.~H. Kang, M.~M.~J. Kabir, Y.~Liu, and R.~Wasinger,
  ``Investigation and improvement of multi-layer perceptron neural networks for
  credit scoring,'' \emph{Expert Systems with Applications}, vol.~42, no.~7,
  pp. 3508--3516, may 2015. [Online]. Available:
  \url{http://dx.doi.org/10.1016/j.eswa.2014.12.006}
\BIBentrySTDinterwordspacing

\bibitem{rossquinlan1993}
J.~R. Quinlan, ``C4.5: Programs for machine learning,'' in \emph{Morgan
  Kaufmann series in machine learning.}, 1993.

\bibitem{quinlan1999simplifying}
------, ``Simplifying decision trees,'' \emph{Int. J. Hum.-Comput. Stud},
  vol.~51, p. 497, 1999.

\bibitem{Lohweg2013}
\BIBentryALTinterwordspacing
V.~Lohweg, J.~L. Hoffmann, H.~D\"{o}rksen, R.~Hildebrand, E.~Gillich,
  J.~Hofmann, and J.~Schaede, ``Banknote authentication with mobile devices,''
  in \emph{Media Watermarking, Security, and Forensics 2013}, A.~M. Alattar,
  N.~D. Memon, and C.~D. Heitzenrater, Eds.\hskip 1em plus 0.5em minus
  0.4em\relax {SPIE}, feb 2013. [Online]. Available:
  \url{http://dx.doi.org/10.1117/12.2001444}
\BIBentrySTDinterwordspacing

\bibitem{morgan:generalization}
N.~Morgan and H.~Bourlard, ``Generalization and parameter estimation in
  feedforward nets: {S}ome experiments,'' International Computer Science
  Institute, Berkeley, CA, Tech. Rep. TR-89-017, 1989.

\end{thebibliography}

\end{document}